\documentclass{nature}
\usepackage{lineno}
\usepackage{url}
\usepackage{graphicx}
\usepackage{subcaption}
\usepackage{amsmath}
\usepackage{arydshln}
\usepackage{multibib}
\newcites{M}{Methods References}

\makeatletter

\let\saved@includegraphics\includegraphics
\AtBeginDocument{\let\includegraphics\saved@includegraphics}

\renewenvironment{figure}[1][tbp]
  {\@float{figure}[#1]}
  {\end@float}

\makeatother

\title{Sympathetic Framing: Evaluating AI Alignment across Sociodemographic Groups}
\author{Haran Shani-Narkiss$^{1,\dagger,*}$, Michael Fire$^{2,\dagger}$ \& Oren Tsur$^{2,\dagger}$}

\begin{document}

\maketitle
\begin{affiliations}
 \item University College London, London, England.
 \item Ben-Gurion University of the Negev, Be'er Sheba, Israel.
\end{affiliations}

\begin{abstract}
Large Language Models (LLMs) increasingly serve as information mediators -- summarizing, interpreting, and shaping the ways people encounter public information. Therefore, it is important to understand whether they perceive subtle emotional framing in ways that reflect human readers. We compare human perceptions of sympathetic framing with the perception of seven leading language models, across major political and geopolitical topics. A large panel of 3,011 UK adults, representing the UK population, was surveyed by a leading polling agency (YouGov). Each participant answered a series of questions regarding a set of news headlines spanning three major geo/political conflicts and campaigns. 
Using 432 binary questions addressing 216 news headlines, we collected more than 200,000 human evaluations and compared them with market-leading LLMs. We find that model–human alignment was very high for the top-performing models, but varied substantially between models (from Spearman correlation of 0.789 for GPT-5.2 to 0.41 for Mistral Large). Importantly, high aggregate alignment does not imply uniform correspondence across populations: even the top-performing model showed significant differences between age, education, first language, political awareness, topic knowledge, and prior views. Alignment also varied across political topics, with some models exhibiting pronounced domain-specific failures. These findings, based on the uniquely large panel of participants of rich and varied demographics, show that even an overall high Model-Human alignment may conceal substantial differences in their adequacy for various demographic groups.

\end{abstract}

\noindent
AI Agents, based on Large Language Models (LLMs), are rapidly becoming indispensable mediators of information, profoundly influencing the ways individuals seek information, consume news\cite{newyorker2026culture} and formulate worldviews\cite{zhang2024benchmarking,joren2025sufficientcontextnewlens,liang2025widespreadadoptionlargelanguage,fortune2025openai}.
Considerable focus was given to the impact of the algorithmic feed on the perception of news and social polarization\cite{kramer2014experimental,bakshy2015fbExposure,guess2021consequences,gonzlez2023segregation,repke2024attention}, on LLMs' hallucinations, bias and fairness\cite{gehman2020realtoxicityprompts,li2023survey,koo2024benchmarking,gallegos2024bias,biever2023chatgpt}, the models' ability to distinguish belief from knowledge and facts\cite{suzgun25}, and models' mis/alignment with human values and goals\cite{ouyang2022training,wei2023jailbroken,gabriel2024ethics,rastogi2026whose,ji-etal-2025-language-models,sun-etal-2025-aligned}.
However, another fundamental concern pertains to their alignment with human comprehension of textual nuances, particularly emotional framing.

Framing, an intrinsic element of compelling writing, layers texts with subtle context, shaping the reader's perception beyond mere factual content\cite{goffman1974frame,entman1993framing}. If LLMs fail to process and convey such nuanced framing accurately, their widespread use for content mediation may impair users' ability to appreciate the richness, context, and intricacies of original texts. The focus on emotional framing, beyond a mere stab into literary devices, is important: the framing of news may influence public opinion\cite{vanDijk1991,clark1992asking,entman1993framing,scheufele2007framing}, shape public discourse\cite{gamson1989news,tsur2015frame}, and ultimately increase social polarization and distrust in social and democratic institutions\cite{guess2021consequences,banks2021polarizedfeeds,lorenz2023systematic}.
Thus, beyond evaluating raw predictive or linguistic performance of an AI agent over common benchmarks\cite{de2023can,laskar2023systematic}, it is critical to ask: 
To what degree is an AI Agent's perception of sympathy and framing aligned with human readers and news consumers? Unfortunately, despite the growing interest in `framing' as a language processing task, the assessment of \emph{alignment} over frames is scarce and based on limited annotation\cite{otmakhova2024media_framing,lior2025wildframe,westwood2025measuring}.

In this work, we present the first comprehensive empirical evaluation of the correspondence between human readers' perceptions of emotional framing and those of an array of LLMs. YouGov, a leading global polling agency, recruited a representative sample ($N=3011$) of the U.K. adult population. Participants answered a series of 72 Yes/No questions (out of 432 total questions across the study) about news headlines covering major political and geopolitical conflicts, yielding 216,792 human evaluations in total. Approximately 500 respondents from diverse demographics answered each question-headline pair. Each of the 432 questions was used to prompt seven different LLMs (Claude Sonnet 4.5, GPT-4, GPT-5.2, Grok 4.1-Fast, Gemini 3-Flash, Mistral Large 2512, and DeepSeek 3.2) 100 times each -- yielding a total of 302,400 AI evaluations (43,200 per model) -- to obtain each model's perception of the framing expressed in each text.

\begin{figure}
    \centering
    \includegraphics[width=\linewidth]{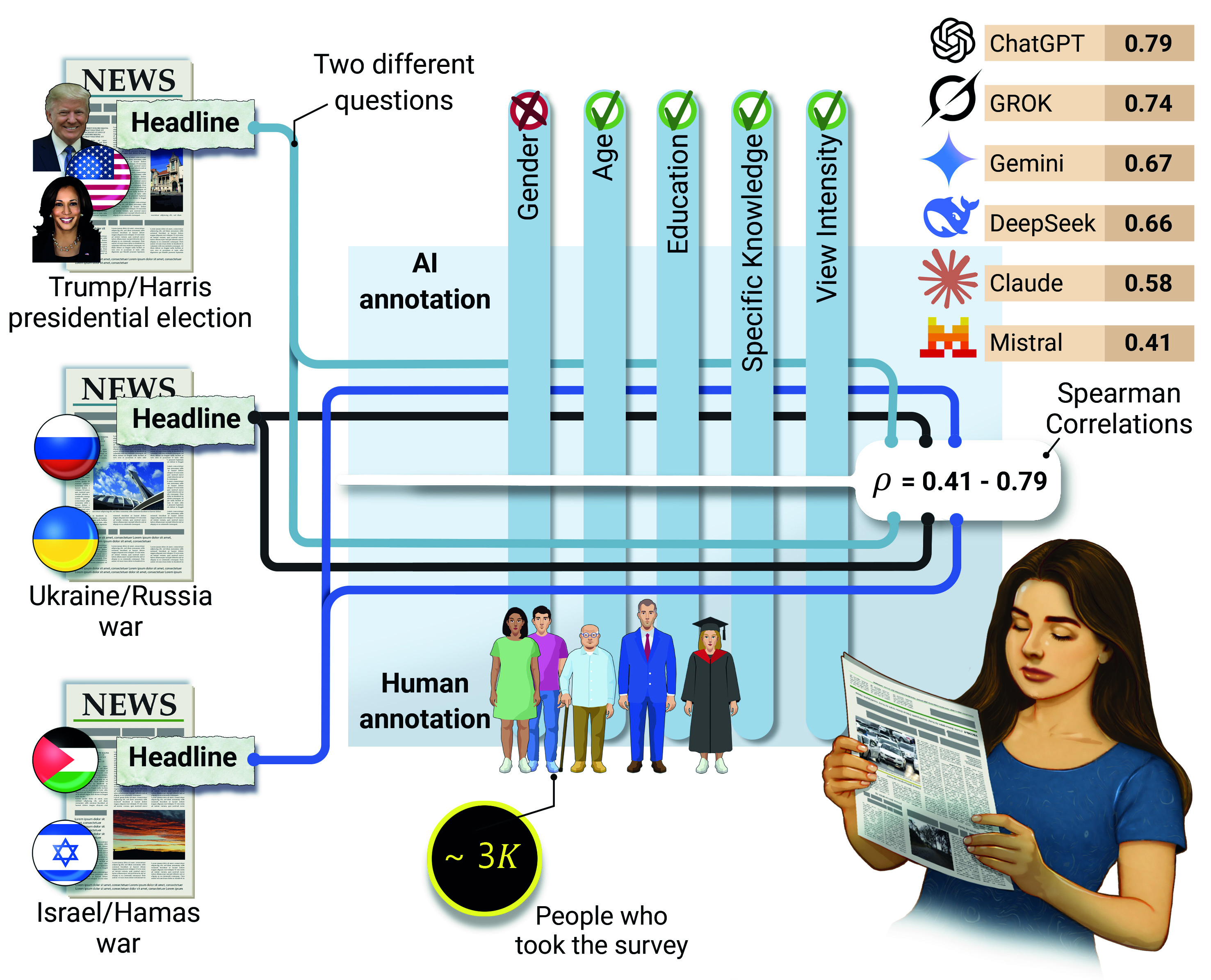}
    \caption{Illustration of the research design and the experimental pipeline. }
    \label{fig:pipeline}
\end{figure}

With hundreds of answers per question, provided by human readers and AI models, we can measure the correlation between human and machine perceptions of emotional framing in news headlines. 
Leveraging the size of the panel, its rich sociodemographic characteristics, and the number of answers per question (see Supplementary Note 1), we were able to correlate and study the alignment between the model's comprehension and the comprehension of fine-grained sociodemographic groups, namely: gender, first language, age, education, NRS (UK-social grade), political awareness, self-reported prior knowledge and preexisting stance related to the different conflicts. An illustration of our experimental pipeline, summarizing the experimental setting, is presented in Fig. \ref{fig:pipeline}.

\section*{Results}
\label{sec:results}

\subsection{Models' Alignment with the General Population.}
The main results are presented in Fig.~\ref{fig:survey_models}. All models showed high and statistically significant alignment with human judgments of emotional framing. However, the level of alignment varied substantially between models and demographic groups. The heat maps (Fig.~\ref{fig:correlations_demographics}a) show a consistent hierarchy of models over demographic and attitudinal subgroups: GPT-5.2 performed best overall, with Spearman’s $\rho = 0.79$ across all respondents, followed by Grok ($\rho = 0.74$) and GPT-4 ($\rho = 0.71$). Gemini and DeepSeek showed intermediate alignment, with overall correlations of $\rho = 0.67$ and $\rho = 0.66$, respectively. Claude achieved weaker alignment ($\rho = 0.58$), and Mistral was the least aligned model ($\rho = 0.41$).

The ordering of the models was broadly stable across demographic subgroups. GPT-5.2 achieved the highest alignment across most demographic and attitudinal categories, typically reaching $\rho = 0.75-0.79$, dropping to 0.7-0.71 for older responders and responders with no formal education. Grok and GPT-4 followed a similar pattern at slightly lower levels, whereas Gemini and DeepSeek generally occupied the middle range. Claude and Mistral showed lower correlations across most subgroups.

\subsection{Alignment over Topics.} Alignment levels also varied across topics. However, some models were more stable than others: the topical alignment levels of GPT-5.2 varied between 0.74 (Trump-Harris) to 0.79 (Russia-Ukrain) while the topical alignment of Grok ranged from 0.62 (Trump-Harris) to 0.77 (Israel-Palestine). The poor overall alignment of Mistral ($\rho=0.41$) are the result of the lack of alignment over Israel-Palestine ($\rho = 0.09$), as it achieved alignment of 0.59 and 0.52 over Russia-Ukrain and Trump-Harris, respectively. Full results by topical breakdown are available in Supplementary Note 5).

\subsection{Alignment Variations across Sub-demographic Groups.} Shifting the focus to the variance in alignment with different sub-demographics, we focus on GPT-5.2, the best performing model (alignment levels are presented in Fig.~\ref{fig:survey_gpt52}; detailed results are available in Tables S2 and S3 in Supplementary Note 4.4). No statistically significant difference was found between male and female respondents ($\rho = 0.784 ~\text{and} ~0.783$, respectively). A statistically significant, though small, difference was found between speakers of English as First Language (L1) and non-native speakers ($\rho_{L1} = 0.790$ vs. $\rho_{L2} = 0.749$; $p=  0.006$).

Age and education showed clearer subgroup differences. The alignment was highest with respondents aged 35–44 ($\rho = 0.790$) and lowest among respondents aged 75+ ($\rho = 0.711$), a significant max–min (range) difference with $p \leq  0.001$). The 18–24 versus 75+ comparison was also significant after correction. For education, alignment was lowest among respondents with no formal qualifications ($\rho = 0.698$) and higher among those with secondary, undergraduate, or postgraduate education ($\rho = 0.781 - 0.790$). The contrast between no formal qualifications and secondary education was significant, as was the contrast between no formal qualifications and Masters/PhD respondents.

\begin{figure}[p]
    \centering

    \begin{subfigure}{0.96\textwidth}
        \centering

        \includegraphics[
            width=\linewidth,
            height=0.22\textheight,
            keepaspectratio
        ]{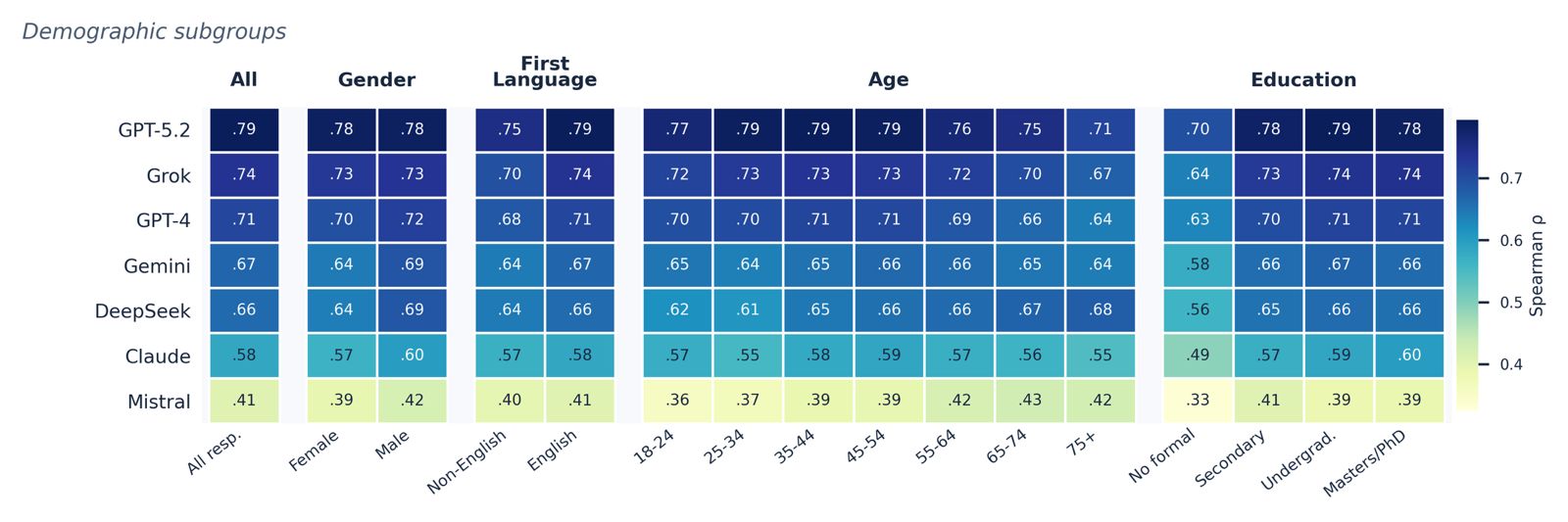}

        \vspace{2mm}

        \includegraphics[
            width=\linewidth,
            height=0.22\textheight,
            keepaspectratio
        ]{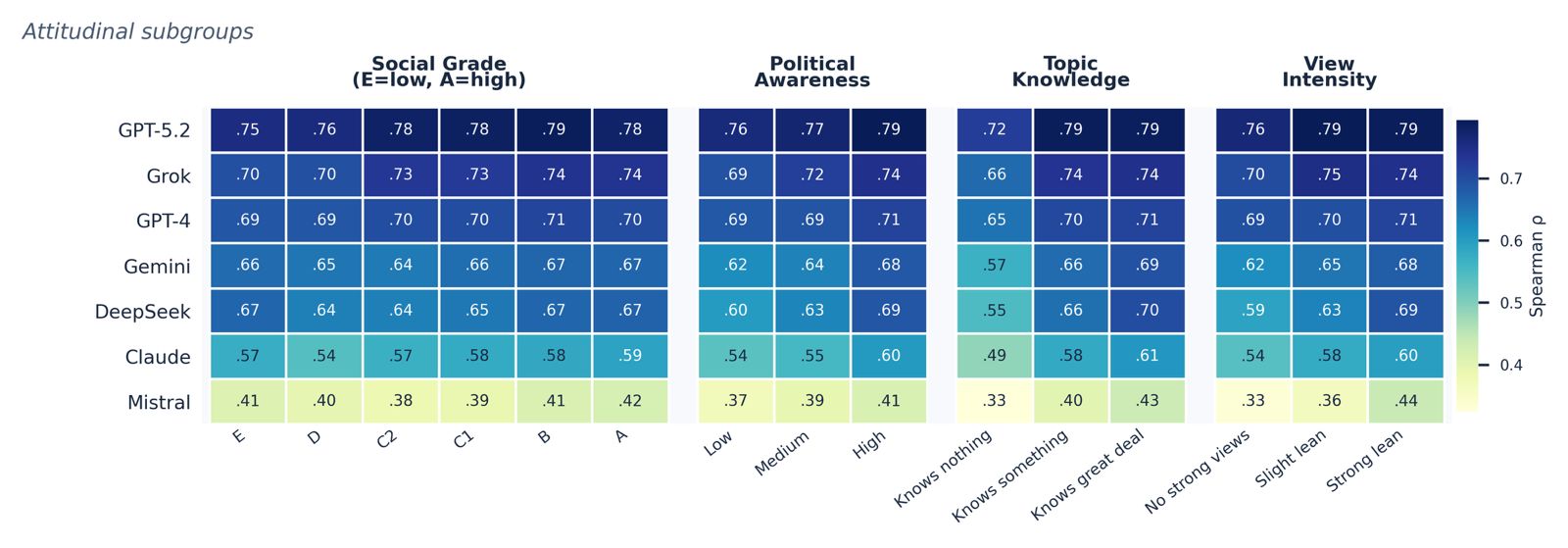}

        \caption{Spearman correlation coefficients between selected
        demographic groups and each model.}
        \label{fig:survey_models}
    \end{subfigure}

    \vspace{3mm}

    \begin{subfigure}{0.96\textwidth}
        \centering

        \includegraphics[
            width=\linewidth,
            height=0.25\textheight,
            keepaspectratio
        ]{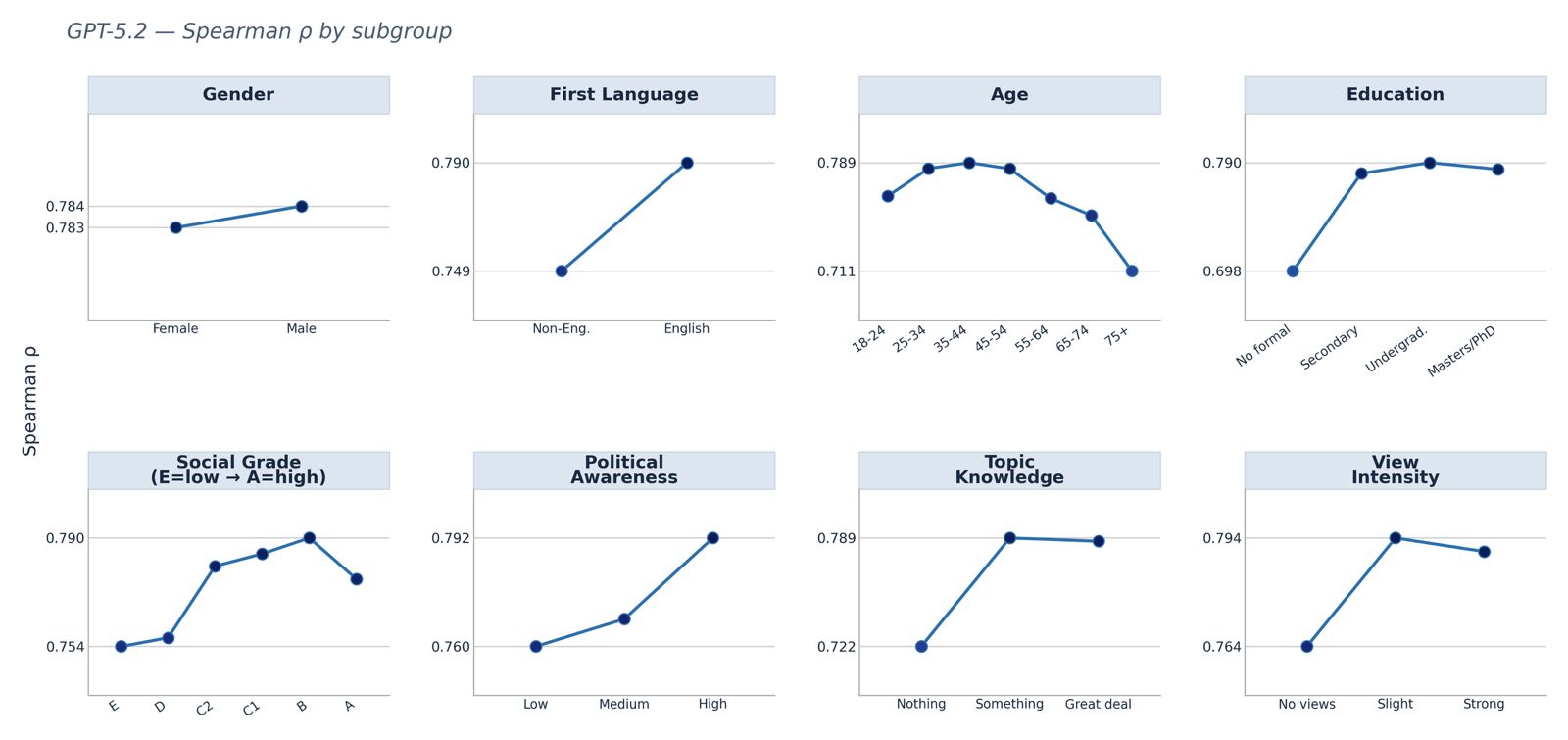}

        \caption{Spearman correlation coefficients between selected
        demographic subgroups for GPT-5.2.}
        \label{fig:survey_gpt52}
    \end{subfigure}

    \caption{\textbf{Alignment of AI models with the overall survey
    population and specific demographic groups.}
    Social-grade categories are defined according to the UK National
    Readership Survey (NRS).}

    \label{fig:correlations_demographics}
\end{figure}

UK-social grade, defined by the National Readership Survey (NRS) is commonly used to assess and explain differences in health, media consumption, and political discourse within the UK population\cite{boykoff2008cultural,kotz2009explaining,yates2018social}. We find that grade exhibit only limited variation in alignment level: the largest difference was between grades B and E ($\rho = 0.790$ versus $\rho = 0.754$; $p \leq 0.05$), while adjacent and extreme comparisons were not significant after correction.

Statistically significant differences in alignment levels between subgroups of different political awareness, knowledge, and attitude (predesposition) were also observed.  GPT-5.2 was more aligned with respondents reporting high political awareness than with those reporting low awareness ($\rho = 0.792$ vs. $\rho = 0.76$; corrected $p \leq 0.001$). Topic-specific knowledge showed a similar pattern: respondents who ``knew nothing'' about the topic had lower alignment ($\rho = 0.722$) than those who know ``something'' or know ``a great deal'' ($\rho = 0.79$). 
Finally, alignment was lower among respondents with no strong views on the conflict ($\rho = 0.764$) than among those with a slight or strong lean ($\rho = 0.79$).

Together, these results show that leading LLMs closely track human judgments of emotional framing across diverse demographic and attitudinal groups. At the same time, alignment is not uniform: it  degrades monotonically with age, reduced over lower levels of education, those with lower political or topic-specific knowledge, and those without strong views. It is also reduced in populations of non-native English speakers.

\section*{Discussion}
\label{sec:discussion}
In the experimental setting described above, we use a number of LLMs as ``readers.'' It is important to distinguish between an LLM's capacity to perceive framing and its ability to generate content reflecting a specific perspective, e.g.,\cite{potter2024hidden,taubenfeld2024systematic}.

Some studies explore how framing is manifested in AI-generated news content, revealing that LLMs tend to exhibit more pronounced framing than human authors, particularly in politically and socially sensitive contexts\cite{pastorino2025frameinframeout}, inline with studies addressing LLM bias\cite{gehman2020realtoxicityprompts,li2023survey,biever2023chatgpt,koo2024benchmarking,gallegos2024bias}. It is also important to distinguish between the use of an LLM as a classifier of explicit forms of sentiment, stance or even emotions, and their performance over more nuanced categories such as framing\cite{ziems24llms4css}.

These results carry significant practical and ethical implications for public opinion research and AI deployment. We find that some models are well aligned with human reading of sympathetic framing of news stories, while other models are poorly aligned. However, while the leading models perform generally well across all demographic subgroups and topics, they do exhibit varying levels of alignment across sociodemographic groups and topics.  
These results highlight the potential and the pitfalls of AI-based surveys to complement classical human-based surveys and and social experiments with synthetic polling \cite{ashokkumar2026large}, offering a path for avoiding these pitfalls. 
These observations offer another perspective on the important discussion about alignment and AI ethics, suggesting that differential alignment protocols -- aligning models differently catering to different demographics and cultural norms -- may be needed.

It is important to note that differential alignment should be considered  with great caution as demographic alignment bears significant ethical implications: Aligning a model inappropriately (e.g., a non-transparent decision on the ``appropriate'' alignment profile made by the model developers)  introduces an inherent yet elusive bias that influences the user's perception of the news. Furthermore, given models' penchant for sycophancy\cite{fanous2025sycevalevaluatingllmsycophancy,ChengEtAl2025} -- generating content that reinforces the user's perceived views\cite{sharma2024towards,cheng2025social}, and the human comfort with echo-chambers, may result in increased polarization and loss of trust\cite{carro2024flatteringdeceiveimpactsycophantic}.

Several methodological nuances and limitations warrant consideration. New LLMs and new versions of existing LLMs are released frequently. At the same time,  results vary across models and versions (e.g., GPT4 and GPT5.2 in Fig.~\ref{fig:correlations_demographics}a). We note that in this work, we introduced a sound methodology that could be applied in testing the alignment with any black-box LLM. Moreover, the unique survey data we release can serve as a benchmark for future studies.

Mistral and DeepSeek exhibit a unique trend as they are better aligned with older demographics. These results, suggesting that different models are based on different practices in training and post-training tuning, yet again highlight the need for transparency and disclosure.  
We note that while interaction effects between demographics and content (e.g., age and education, or topic awareness) are probably at play and are not addressed in this work, the reported correlations are sufficient to demonstrate significant differences in alignment  between models and demographic groups. 

Finally, the analysis and insights offered in this manuscript are possible due to the uniquely large representative panel we have curated. Although focused on the UK population, this dataset, including the rich socio-demographic breakdown and the participant evaluations of the survey items, can serve as an important benchmark and a blueprint for future research of alignment and the utility of LLMs and AI agents serving as information mediators.


\bibliography{llm_representation}

\begin{methods}
\label{sec:method}

\paragraph{Data.} News headlines were extracted from the GDELT Project\citeM{leetaru2013gdelt}. GDELT is a project supported by Google Jigsaw, monitoring and curating ``the world's broadcast, print, and web news,'' using a set of predefined keywords related to three major political and geopolitical conflicts: The ongoing Russo-Ukrainian war (2022--), the war in Gaza (2023-2025), and the 2024 U.S. Presidential Campaign. The initial dataset contained $\sim 800K$ unique headlines, from which a subset of headlines was retained: for each topic $k=36$ headlines were sampled at random, and another $k'=36$ headlines, deemed ``nuanced'' (framing), were selected. The final dataset consists of 216 headlines (72 per conflict). The exact procedure used for headline selection is available in Supplementary Note 1, along with the final list of headlines.

\paragraph{Experimental design.} Our experimental design involves the use of two mirrored yes/no questions  to assess the sympathetic framing of a given headline:

\begin{enumerate}
    \item \texttt{Does this text create sympathy towards A? <TEXT>}
    \vspace{-12pt}
    \item \texttt{Does this text create sympathy towards B?  <TEXT>}
\end{enumerate}

Combining the binary answers for the mirrored questions, the evaluation of a \emph{headline} can take four values: (i) sympathy for neither, (ii) sympathy for A, (iii) sympathy for B, or (iv) sympathy for both A and B.

While a single evaluation for a single headline–question pair may offer limited signal, aggregating multiple evaluations provides a robust empirical measure. Soliciting $n$ binary (Yes/No) responses to a specific question $q_i$, we use $r_i \in [0,1]$ to denote the proportion of positive responses out of $n$. Thus, $r_i$ quantifies the level of sympathy toward the target entity (A or B). Although $r_i$ could be binarized using a threshold ($r_i < 0.5$ versus $r_i \ge 0.5$) to form a Bernoulli trial, retaining the continuous raw proportion that better reflects the subtlety, opaqueness, or explicitness of the perceived framing.

Given a set of questions $Q = \{q_1, \dots, q_k\}$ and a model $M$, we obtain two corresponding vectors: $R^H = \{r^H_1, \dots, r^H_k\}$ for Human responses and $R^M = \{r^M_1, \dots, r^M_k\}$ for Model outputs. Here, $r_i^H$ denotes the proportion of positive human responses to $q_i$, while $r_i^M$ denotes the proportion of positive responses obtained by prompting model $M$, $n=100$ times with question $q_i$. This repeated prompting protocol, commonly referred to as self-consistency testing, is used to account for the stochastic decoding inherent to large language models\citeM{wangself}.

\paragraph{Measuring Alignment.} Having two corresponding lists of values, $R^H$ and $R^M$, we use the Spearman correlation $\rho(R^H,R^M)$ to quantify the alignment between the human perception and the model's perception of news articles. Given the size and the representative nature of panel, we can correlate the model's perception with the perception of a specific demographic $\Lambda$ by considering only responses provided by participants of that demographic. The alignment between this subgroup and a model is given by $\rho(R^{H_\Lambda},R^M)$. Spearman’s rank correlation coefficient, $\rho$, ranges from $-1$, indicating a perfect inverse monotonic association, to $+1$, indicating a perfect positive monotonic association, with $0$ indicating no association. 
Following common interpretive conventions, correlations around 0.4 are typically considered moderate, whereas correlations near 0.8 indicate strong to very strong alignment\citeM{evans1996straightforward,schober2018correlation,akoglu2018user}. 
We test the significance of the results via permutation testing\citeM{tibshirani1993introduction,good2013permutation,holt2023permutation}, see formal definitions, implementation details, and worked examples in Supplementary Note 4. 

\paragraph{Survey.}  A panel of participants was recruited by YouGov. Candidates were given a short explanation about the nature of the survey, after which they could decline participation or opt-in. The process conformed to YouGov's Code of Conduct \& Ethics\citeM{youGovEthics} and was approved by an Institutional Review Board (\#SISE-2024-48). 
Opting-in, participants first answered questions about prior knowledge (``How much, if anything, would you say you know about each of the following topics?'') and about their preexisting stance (``Would you say you have a more favorable view of Side A or Side B, or neither?''). Participants were randomly assigned to one of six experimental groups (splits), each group was exposed to 36 headlines (12 per topic), for which they were asked to answer the twin yes/no questions described above.
The recruited panel ($N=3011$) formed a representative sample of the adult UK population, providing rich demographic data, including self-reported topic knowledge and emotional attachment. 
Full details about the recruitment process and the demographic distribution of responders (overall and within each split) are available in Supplementary Note 2. Headline assignment to splits can be found in Supplementary Note 1. Details about the administration of the survey are available in Supplementary Note 3.

\end{methods}


\bibliographystyleM{naturemag}
\bibliographyM{llm_representation}

\section*{Data Availability}
All anonymized human survey data ($N=3,011$) and LLM evaluation outputs ($302,400$ responses) will be fully available upon publication.

\section*{Code Availability}
All custom Python scripts for prompting, evaluation, and statistical analysis will be accessible upon publication on GitHub.

\section*{Acknowledgements}
This work was supported by [Grant Agency / Funding Body] under Grant No.~\mbox{[XXXXXX]}, with computational infrastructure provided by [Institution / Cloud Provider]. We thank [Name / Research Group] for insightful comments on earlier versions of this manuscript. Survey panel recruitment and administration were conducted via YouGov UK. Ethics approval for human subject data collection was granted by the Ben-Gurion University of the Negev Institutional Review Board (IRB~\#SISE-2024-48).



\section*{Additional Information}
Supplementary Information is available for this paper.
Correspondence and requests for materials should be addressed to the authors.


\end{document}



\section{Survey Questions}
\label{sm:questions}

\paragraph{Headline Retrieval from GDELT.}
The Global Database of Events, Language, and Tone (GDELT)\cite{leetaru2013gdelt} is a comprehensive open-access project that continuously monitors global news media outlets, automatically extracting and codifying reports of events, key actors, locations, and the tone of coverage in over 100 languages\cite{leetaru2013gdelt}. By providing structured data on a myriad of socio-political events and related news content, GDELT enables large-scale, quantitative studies of worldwide media trends.

A fundamental analytic technique for exploring GDELT data is keyword search. Researchers can use targeted keywords, phrases, or Boolean expressions to identify and isolate news articles, events, or themes of interest within the dataset. This facilitates topic-based filtering, temporal or geographic trend mapping, and content analysis at an unprecedented scale. The unique coverage supports research event detection, sentiment analysis in multilingual news media, the spread of misinformation, analyzing crisis communication, news framing, climate change recognition, and modeling information diffusion, e.g., 
\cite{ward2013comparing,kwak2014first,imran2015processing,raleigh2023political,mudassar2025global}.

\paragraph{Data Collection.}
News headlines were aggregated from a range of mainstream outlets via GDELT feeds and curated repositories to ensure broad representation of political leanings and rhetorical styles. Initial data collection focused on three subjects: (1) the 2024 U.S. Presidential race, (2) the Israel-Gaza war (2023), and (3) the Russia-Ukraine war. For each subject, headlines were collected within specific temporal windows (U.S. Presidential race: 15/6/2024–15/9/2024; Israel-Gaza war: 7/10/2023–7/2/2024; Russia-Ukraine war: 24/2/2022–24/6/2022) using relevant keywords or letter sequences (U.S. Presidential race: “Kamala”, “Harris”, “Trump”; Israel-Gaza war: “Israel”, “Gaza”, “Palestin”, “Antisem”, “Hamas”, “IDF”; Russia-Ukraine war: “Russia”, “Ukraine”, “Kyiv”, “Kiev”, “Zelenskyy”, “Donbas”, “Luhansk”, “Putin”). This process produced 1,672,909 headlines (U.S. Presidential race: 414,379; Israel-Gaza war: 713,801; Russia-Ukraine war: 544,729). After removing duplicates, the dataset comprised 824,876 unique headlines (U.S. Presidential race: 199,728; Israel-Gaza war: 343,322; Russia-Ukraine war: 281,826).

\paragraph{Survey design and headline selection.}
The survey was designed to balance topic coverage and cognitive load, allowing sound statistical analysis while maintaining a reasonable completion time that caps cognitive load and ensures response quality. 
To approximate real-world exposure, we included a set of randomly sampled news headlines. However, since most headlines in the source datasets were either strictly factual (e.g., "Russia-Ukraine war: What happened today") or easily decided toward one side (e.g., "Fears for Mariupol defenders after surrender to Russia"), we supplemented these with a curated set of more nuanced and challenging items (e.g., "Russia rules out any threat to Ukrainian civilians during military action"). This approach enabled the inclusion of less trivial cases while preserving representativeness.

The two-stage headline selection was done as follows: First, 108 headlines were randomly sampled from the compiled datasets. Second, to identify nuanced items, the full dataset was processed using GPT-4 with the standard prompt to assess whether headlines elicited sympathy towards one or more sides of a conflict. Most headlines did not elicit sympathy; among those who did ($\sim 25\%$), the vast majority ($\geq 95\%$) elicited sympathy toward one side only.
To enrich the dataset with more complex cases, we implemented a targeted selection procedure. Headlines that produced at least one positive response in the initial screening were retained and re-evaluated across 10 independent runs using the same model and prompt. From this subset, 108 headlines were manually selected based on either (i) consistent elicitation of sympathy toward both sides or (ii) variability in responses across the ten runs.
The final corpus consisted of 216 headlines: 108 randomly sampled and 108 selectively curated, a total of 72 headlines per domain (36 random, 36 selected). In addition, four control headlines per domain (two with overtly sympathetic framing and two completely unrelated to the topic, see Fig.~\ref{fig:control_q}) were included to support participant quality control. These control items were used by YouGov as an exclusion criteria.

All headlines were manually verified for authenticity and edited to remove problematic features (e.g., non-English characters, mixed topics), matching them with the original source.

An overview of the selection procedure is provided in Table \ref{tab:data_collection_headlines}. The final instrument comprised 432 questions based on 216 headlines. Participants were assigned to one of six groups (\emph{Splits}), each of which evaluated a batch of 36 headlines (72 questions). The full set of headlines, including source information and data splits, is available in an external repository \href{https://docs.google.com/spreadsheets/d/1HU4BSkdFZ4GUfE5FQc92GYJYYmX-hSMbL97LKLO7d10/edit?usp=sharing}{(Google Sheet)}.

\begin{table}[ht]
    \centering
    \footnotesize
    \begin{tabular}{p{2.5cm}|p{2cm}|p{3.5cm}|p{1.8cm}|p{1.5cm}|p{1.5cm}}
        \textbf{Topic} & \textbf{Time Span} & \textbf{Keywords} & \textbf{\# matches} & \textbf{Unique} & \textbf{Final set} \\ \hline
        
        U.S. 2024 Presidential race & 15/6/2024 -- 15/9/2024 & Kamala, Harris, Trump & 414,379 & 199,728 & 72 \\ 
        \hline
        Israel-Gaza war & 07/10/2023 -- 07/02/2024 & Israel, Gaza, Palestin, Antisem, Hamas, IDF & 713,801 & 343,322 & 72 \\ \hline
        Russo-Ukraine war & 24/02/2022 -- 24/06/2022 & Russia, Ukraine, Kyiv, Kiev, Zelenskyy, Donbas, Luhansk, Putin & 544,729 & 281,826 & 72 \\
    \end{tabular}
    \caption{Overview of the headlines selection process. The Final Set includes 36 randomly sampled headlines per topic and 36 headlines sampled from the set of ``nuanced'' headlines.}
    \label{tab:data_collection_headlines}
\end{table}

\section{Panel Demography}
\label{sm:demography}
\begin{figure}[H]
    \centering
    \includegraphics[width=\linewidth]{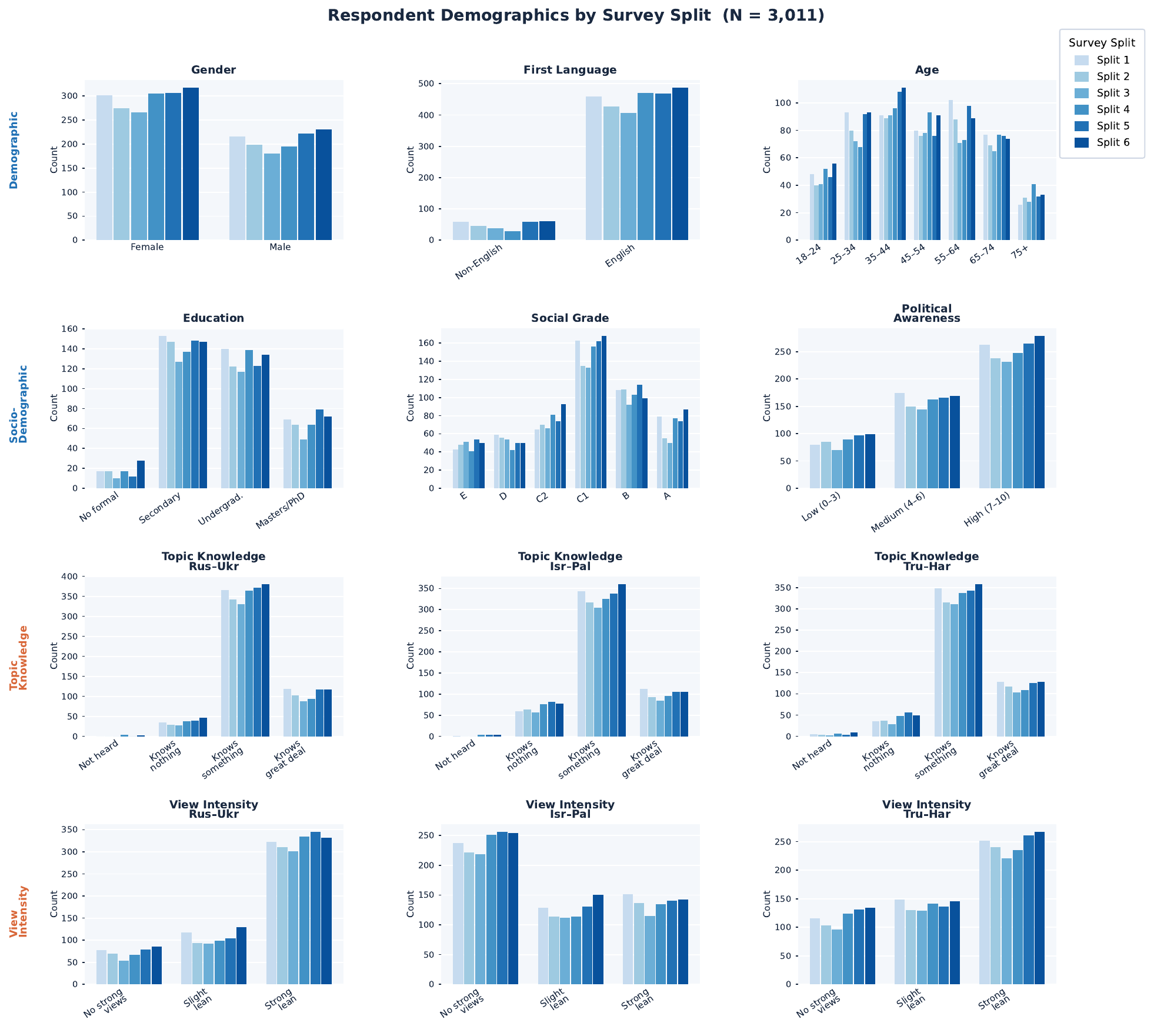}
    \caption{Demographic distributions of the study sample. Subplots show (A) Gender, (B) Education, and (C) Age group classifications.} 
    \label{fig:demographics}
\end{figure}

A representative sample of the U.K. adult population was recruited by YouGov, a leading global polling agency. Candidates were given a brief explanation of the survey's nature, after which they could decline to participate or opt in. The process conformed to YouGov's Code of Conduct \& Ethics \cite{youGovEthics} and was approved by an Institutional Review Board (\#SISE-2024-48).
Some of the responders ($m=1,308$) opted out, did not complete the survey, or proved inattentive (failed attention questions  that popped up during the survey, see details in \ref{sm:survey_administration}), and were consequently excluded by YouGov’s. In total, valid responses were collected from $n=3,011$ responders. 
The demographic breakdown of this group is provided below.

\section{Survey Administration}
\label{sm:survey_administration}

\begin{figure}[htp]
    \begin{subfigure}{\textwidth}
    \includegraphics[width=\textwidth]{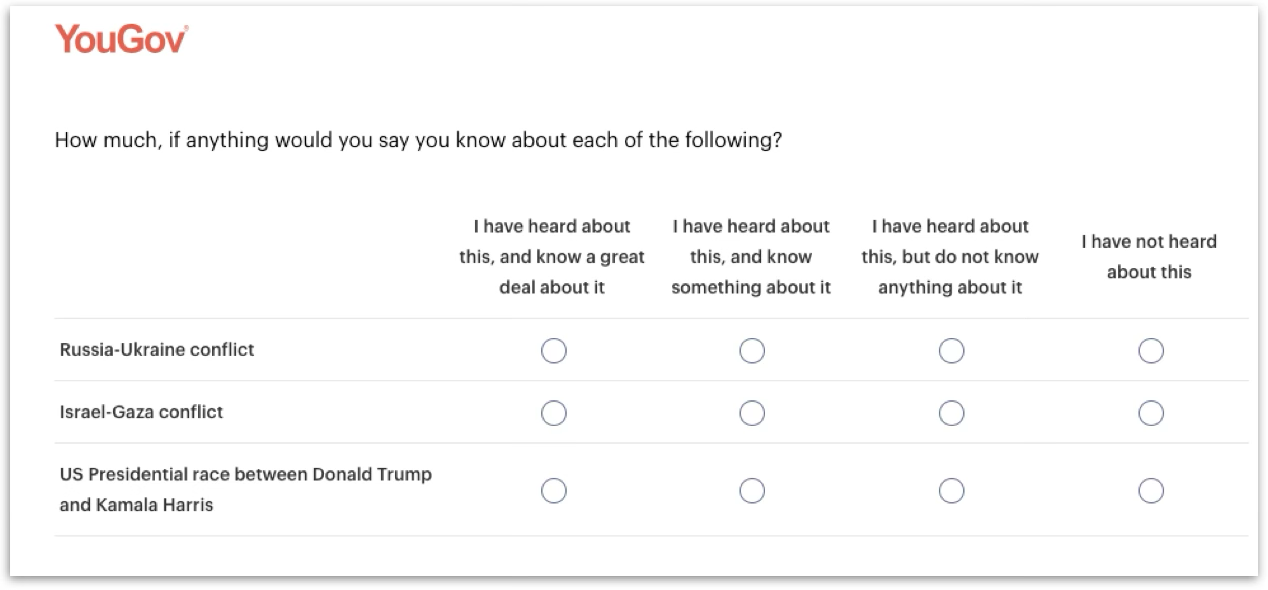}
    \caption{Participants mark their self-assessed knowledge of the topics they will be asked about.}
    \label{fig:self_knowledge}
    \end{subfigure} \hfill
    
    \bigskip 
    \begin{subfigure}{\textwidth}
       \includegraphics[width=\textwidth]{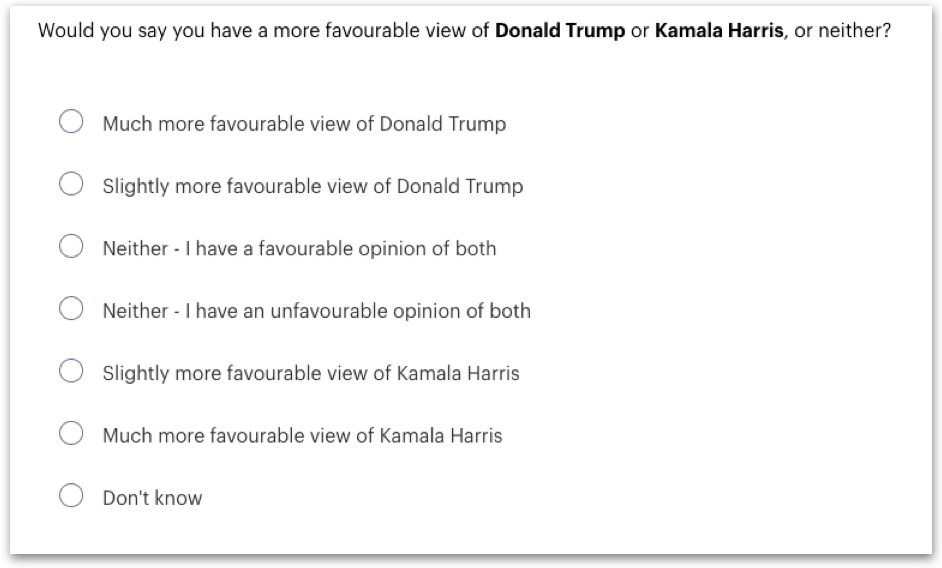}
    \caption{Participants indicate their predisposition regarding the parties involved in the questions they will be presented with.}
    \label{fig:us_fav}
    \end{subfigure}
   
    \caption{Knowledge of the topics and predisposition towards the parties involved}
    \label{fig:survey_screens_prior_knowlwdge_and_predisposition}
\end{figure}

Participants completed the survey through a web application maintained and administered by YouGov. Upon opting in, participants were randomly allocated to one of the six groups and first answered general questions about their knowledge of the topics and their predisposition towards the relevant parties (see Figs.~\ref{fig:self_knowledge} and \ref{fig:us_fav}).  

\begin{figure}[htp]
\center
    \begin{subfigure}{\textwidth}
    \centering
    \includegraphics[width=0.7\textwidth,height=6cm]{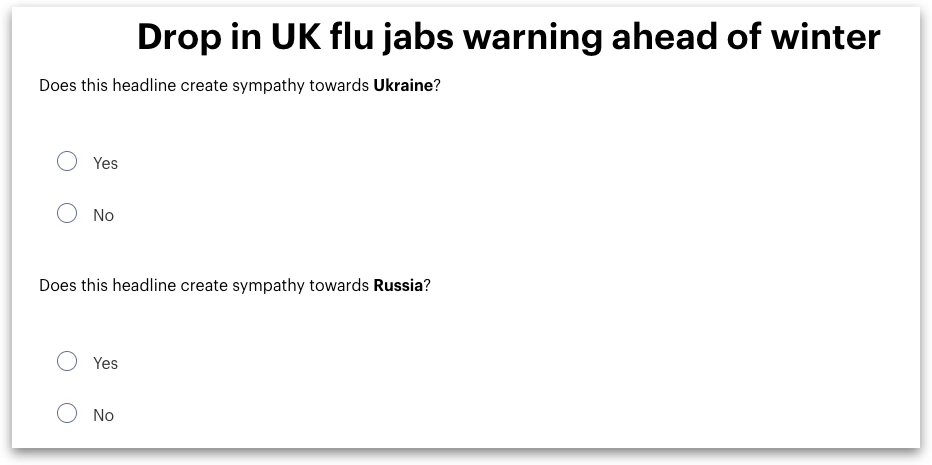}
    \caption{An example of a control question - the text is unrelated to the questions. Participants are expected to answer \texttt{No} on both questions.}
    \label{fig:control_q}
    \end{subfigure} \hfill
    
    \bigskip 
    \begin{subfigure}{\textwidth}
    \centering
       \includegraphics[width=0.7\textwidth,height=6cm]{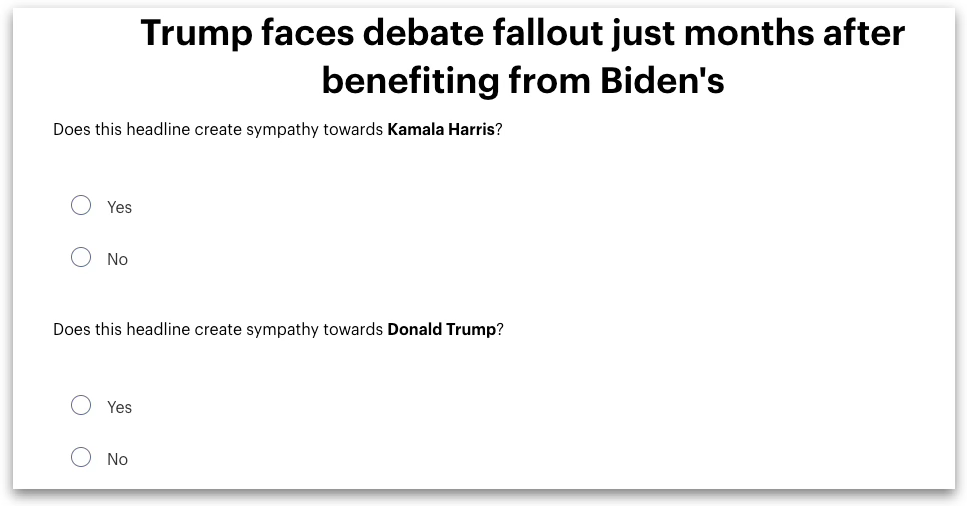}
    \caption{Question about a headline reporting on the Presidential debate of September 10th 2024.}
    \label{fig:debate}
    \end{subfigure}

    \bigskip 
    \begin{subfigure}{\textwidth}
    \centering
\includegraphics[width=0.7\textwidth,height=6cm]{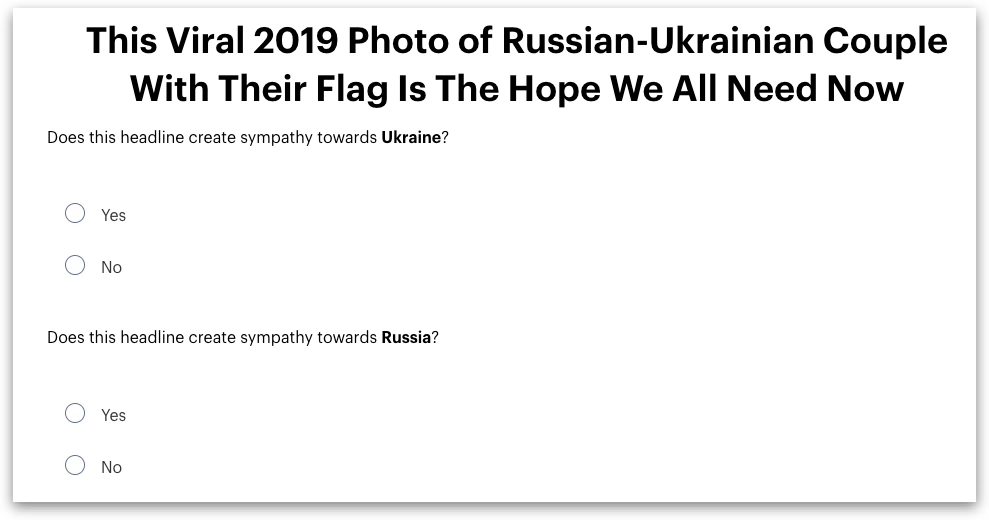}
    \caption{Question about a headline evoking sympathy to both sides in the Russia-Ukraine war. }
    \label{fig:couple}
    \end{subfigure}
    \caption{Examples of survey questions.}
    \label{fig:survey_screens}
\end{figure}

To ensure quality, each survey included a control question placed among the other questions (see Fig.~\ref{fig:control_q}). The answers to these questions were not processed with the other questions. Rather, participants who failed to answer the control questions correctly were disqualified, and their answers were ignored.  

An example of a headline regarding the Presidential campaign is presented in Fig.~\ref{fig:debate}, and an example of a headline potentially invoking sympathy for both sides of the Russo-Ukraine war is presented in Fig.~\ref{fig:couple}.

\section{Significance Testing}
\label{sm:permutation_tests}
The main contribution of this work is the evaluation of the alignment of various AI models with specific demographics. 
Given two corresponding lists of values, $R^H$ and $R^M$, we use the Spearman correlation $\rho(R^H,R^M)$ to quantify the alignment between human and model perceptions of news articles. Given the size and representative nature of the panel, we can correlate the  model's perception with the perception of a specific demographic $\Lambda$ by considering only the responses provided by participants of that demographic. 
The alignment of a demographic subgroup $\Lambda$ with model $M$, is the Spearman rank correlation:

\[
\rho_\Lambda = \rho_\text{Spearman}(R^\Lambda, R^M)
\]

Considering two mutually exclusive demographic subgroups $\Lambda$ and $\Psi$, we denote the difference between the level of alignment of a model with demographic $\Lambda$ and its alignment with demographic $\Psi$ by:
 \[\Delta_{\Lambda,\Psi} = |\rho_\Lambda - \rho_\Psi|\]

We use permutation testing \cite{tibshirani1993introduction,good2013permutation,holt2023permutation} to test whether $\Delta_{\Lambda,\Psi}$ is statistically significant. 

In the remainder of this section, we provide the formal definitions and specify the different realizations of the permutation testing. 

We begin by introducing the formal notation and the general form of the permutation test, followed by details on the specific realization used with respect to different sub-demographic breakdowns: binary, ordinal, and categorical. A worked example is offered with each type of realization. We conclude this Appendix with tables that provide comprehensive descriptions and results for all significance tests of differences in alignment levels between ChatGPT5.2 (the highest overall alignment) and different sub-demographics.

\subsection{Formal Notation}

\paragraph{Questions and Splits.} Let $\{q_i\}_{i=1}^k = Q$ a set of $k$ yes/no questions ($k=432$).

The questions in $Q$ are partitioned into $S$ disjoint subsets (\emph{Splits}) of equal size (in practice, $|S|=6$, see Survey design and headline selection in \ref{sm:questions}). $Q^s$ denotes a specific subset of questions included in the $s\text{-th}$ split ($s \in S$), and use $q^s_i$ if $q_i \in Q^s$.

\paragraph{Responders, Demograpics, and Splits.} Let $U = \{u_j\}_{j=1}^n$ be a group of $n$ responders ($n=3011$).  Each responder is assigned a split $s$ and only answers questions in $Q^s$. We use $u^s_j$ to denote the responder's assigned split. Responders can also be characterized by demographic traits. We use $u^\Lambda_j$ to denote that $u_j$ belongs to a specific sub-demographic $\Lambda$. We use $u^{s,\Lambda}_j$ to indicate the responder's split \emph{and} demographic group. Similarly, we use $U^{s,\Lambda}$ to indicate all responders of demographic $\Lambda$ in split $s$.

Taking \emph{all} responders into account, each question $q_i$ was answered by $\frac{n}{S}$ responders. However, when considering only responders in specific demographic $\Lambda$, questions across different splits may garner a different number of answers due to the distribution of demographic traits and the random assignment of responders to splits. 
For example, considering only Male responders ($\Lambda = \text{Male}$) it may be the case that  $|U^{s,\Lambda}| \neq |U^{s',\Lambda}|$, thus a question $q \in Q^s$ gets a different number of answers by male responders than a question $q' \in Q^{s'}$. 

\paragraph{Survey results and Positive response ratio.} We use $R^H = r^H_1,...,r^H_k$ to indicate the list of ratios of positive answers received for question $q_i$. That is, if $q_i$ is answered by $m_i$ responders, out of which $y_i \leq m_i$ answered positively, $r_i = \frac{y_i}{m_i}$. 
Considering only responders of a specific demographic $\Lambda$, we define $R^\Lambda = r^\Lambda_1,...,r^\Lambda_k$, where $M^\Lambda_i$ and $y^\Lambda_i$ are the number of responders and positive responses of members of $\Lambda$ to $q_i$. Given that $q_i \in Q^s$ we have $r^\Lambda_i = \frac{y\Lambda_i}{m^\Lambda_i}$ and $m^\Lambda_i = |U^{s,\Lambda}|$.

$R^M = \{r^M_i,...,r^M_k\}$ denotes the positive ratio returned by model $M$ to $q_1,..., q_k$. Given each prompt is issued $n$ times, $r^M_i$ is the number of positive answers returned to $q_i$, divided by $n$. 

\paragraph{Measuring Alignment.} The alignment of a demographic subgroup $\Lambda$ with model $M$, is the Spearman rank correlation:

\[
\rho_\Lambda = \rho_\text{Spearman}(R^\Lambda, R^M)
\]

 Given two mutually exclusive demographic subgroups $\Lambda$ and $\Psi$, we define 
 \[\Delta_{\Lambda,\Psi} = |\rho_\Lambda - \rho_\Psi|\]
 to indicate the absolute difference between a models's alignment with demographic $\Lambda$ and demographic $\Psi$. We are interested in testing whether $\Delta_{\Lambda,\Psi}$ is significant. Significance is evaluated through permutation testing.

\paragraph{Permutation testing -- Design Principles.} Three principles apply to every realization of a permutation test used in this paper.

\begin{enumerate}
\item \textbf{Per-participant relabelling:} The unit being re-labelled is the respondent $u_j$, rather than an isolated response to a question $q_i$; under any permutation $\pi$, all $|Q^s| =72$ responses provided by $u^s_j$ are re-assigned together. This preserves within-respondent answer dependence and is therefore conservative relative to a per-question shuffle.

\item \textbf{Stratification by Split:} Permutations are constrained to preserve, within every Split $s \in \{1,\dots,S\}$, the number of respondents at each subgroup level. This matches the experimental design (Split is fixed by the protocol; only the human group label is exchangeable conditional on Split) and rules out chance Split\,$\times$\,label imbalance.

\item \textbf{Fixed $R^M$ vector:} $R^M$ is held fixed across all permutations. The test asks only whether the human side could have produced the observed alignment under random labeling; the model provides the reference axis.
\end{enumerate}

\paragraph{Permutation testing -- General form.} Given a set $\Gamma$ of mutually exclusive demographic sub/groups, let $\pi$ be a random partitioning function defined as:

\begin{equation}
\pi: U \rightarrow \{\overline{U^{s,\gamma}}\}_{s,\gamma \in S \times \Gamma}
\label{eq:statified_permutation}
\end{equation}

S.t. the design principles stated above: 

\[\bigcup_{s,\gamma \in S \times \Gamma}  \overline{U^{s,\gamma}} = U \text{~;~}
\bigcap_{s,\gamma \in S \times \Gamma}  \overline{U^{s,\gamma}} = \phi \text{~; and ~}
|\overline{U^{s,\gamma}}| = |U^{s,\gamma}| ~~ \forall_{s,\gamma \in S \times \Gamma}
\]

\vspace{10pt}
For any two sub-groups $\Lambda, \Psi \in \Gamma$, we can now compute $\overline{R^\Lambda}$ and $\overline{R^\Psi}$,  find $\overline{\rho_\Lambda}$ and $\overline{\rho_\Psi}$, and define the absulute difference: 
\[\overline{\Delta_{\Lambda,\Psi}} = |\overline{\rho_\Lambda} - \overline{\rho_\Psi}| \]

The two-tailed $p$-value is therefore given by:

\begin{equation}
p_{\Lambda,\Psi} = \frac{\sum_1^z \mathrm{I}(\Lambda,\Psi, \pi)}{z}
\label{eq:p_raw}
\end{equation}

where $z$ is the number of permutations (we use $z=10^5$) and 

\begin{equation}
\mathrm{I}(\Lambda,\Psi, \pi) = 
\begin{cases}
1 & \text{if } \Delta_{\Lambda,\Psi} \leq \overline{\Delta_{\Lambda,\Psi}} \\
0 & \text{otherwise}
\end{cases}
\label{eq:I}
\end{equation}

\subsection{Binary Attributes (Gender, First Language)}
\label{sec:perm-binary}

The respondents included in the panel gathered by YouGov reported gender in a binary way: \emph{Male} and \emph{Female}. Similarly, since all texts were taken from English media outlets and the survey was administered in English, we treat native language as a binary variable, taking the values \emph{English} and \emph{non-English}. 

In the binary case, the permutation test is realized in a straightforward way -- the two-tailed $p$-value is given by the general form in Equation \ref{eq:p_raw}. However, in the binary case, the two sub-demographics fully cover $\Gamma$:

\[
p_{\Lambda,\Psi} = \frac{\sum_1^z \mathrm{I}(\Lambda,\Psi, \pi)}{z}
\]

\begin{examplebox}{Worked Example --- Binary Case: Gender (Female vs.\ Male)}

\textbf{Setup:} The Gender attribute has two groups. Group $\Lambda$ = Female ($|U^\Lambda| = 1,769$) and group $\Psi$ = Male ($|U^\Psi| = 1,242$), for a pooled sample of $|U^\Lambda \bigcup U^\Psi | = 3,011$ respondents. The per-Split cell counts are:

\begin{center}
\small
\begin{tabular}{cccc}
\toprule
Split & Split size & Female ($\Lambda$) & Male ($\Psi$) \\
\midrule
1 & 517 & 301 & 216 \\
2 & 473 & 274 & 199 \\
3 & 446 & 266 & 180 \\
4 & 500 & 305 & 195 \\
5 & 528 & 306 & 222 \\
6 & 547 & 317 & 230 \\
\bottomrule
\end{tabular}
\end{center}

\textbf{Observed statistics.} Computing yes-rate vectors $R^\text{Female}$ and $R^\text{Male}$ over $k = 432$ questions and correlating each with $R^M$ (in this and subsequent examples $M$ is ChatGPT5.2):
\begin{equation*}
  \rho_{Female} = 0.7834, \quad \rho_{Male} = 0.7841, \quad
  \Delta_{Female,Male} = \rho_{\text{Female}} - \rho_{\text{Male}} = -0.0007.
\end{equation*}

\textbf{One permutation step (illustration):} In Split~1 we have 301 Females and 216 Males. We draw a random permutation of 517 labels (301 F's and 216 M's) within Split~1 independently, and similarly for each of the other five Splits. The per-Split $\frac{F}{M}$ counts are identical to the observed counts -- only the \emph{identities} of which respondents are labelled Female vs.\ Male may change. Thus, if $u_i$, originally in $U^{s=1,Female}$ was reassigned by $\pi$ to $\overline{U^{s=1,Male}}$, all of her $72$ answers are now considered to be provided by a male responder.
\\

\textbf{Result after $z = 10^{5}$ permutations:} $90,238$ permutations produced $\mathrm{I}(\circ)=1$, 
giving
\begin{equation*}
  p_{Female,Male} = \frac{90,238}{100,000} \approx 0.902 \quad (\text{ns}).
\end{equation*}

{\bf Interpretation:} Both groups align equally with GPT-5.2; there is no statistically significant gender gap.
\end{examplebox}


\subsubsection{Multi-Category \emph{Ordinal} Attributes (Age, Education, Social Grade, General Political Awareness)}
\label{sec:perm-ordinal}

Some demographic attributes have ordinal sub-categories. Typical examples are age and education. Similarly, self reported levels of general political awareness, the level of knowledge of a specific conflict or the predisposition towards a specific side in a specific conflict are ordinal (see Fig.~\ref{fig:survey_screens_prior_knowlwdge_and_predisposition} in Appendix \ref{sm:demography}).


\paragraph{Family of named pair comparisons} Given $c$ ordinal sub-categories of a demographic trait $\Gamma$ ($|\Gamma| = c \geq 3$), we test the significance of $ \Delta_{\Lambda,\Psi}$ in a family of $c+1$ configurations of $(\Lambda,\Psi)$ pairs \cite{hochberg1987multiple}:

\vspace{10pt}
$(\Lambda=i, \Psi=i+1) ~~\forall i\in \{1,...,c-1\}$ \hfill $c-1$ ordinal  \emph{adjacent} pairs \\

\vspace{-3pt}
$(\Lambda=1, \Psi=c)$ \hfill the two sub-categories at the \emph{extremes}
\[\hspace{14pt}(\Lambda = \arg\min_{i}\rho_i, \Psi = \arg\max_{i}\rho_i) ~~~~~~~~ \text{sub-categories of \emph{min/max} alignment}\]

\paragraph{Marginal and family-wise $p$-values:}
The marginal (raw) $p$-value for pair $(\Lambda,\Psi)$ is given by the general form in Equation \ref{eq:p_raw}. However, the \emph{adjacent} and \emph{extreme} pairs share respondents and therefore marginal $p$-values are dependent Bonferroni and Benjamini-Hochberg corrections are suboptimal. We thus use the exact \emph{maximum-statistic} family-wise null distribution: for \emph{each} permutation $\pi$ and adjacent pair of sub-categories $\Lambda$ and $\Psi$ we replace $\overline{\Delta_{\Lambda,\Psi}}$ in Equation \ref{eq:I}, with $\overline{\Delta_{\gamma,\psi}}$, where 
\[\gamma = \arg\min_{i}\rho_i ~\text{and}~  \psi = \arg\max_{i}\rho_i.\]
The $p$-value based on this maximal Family-Wise Error Rate statistic is denoted by: 

\hspace{30pt} $p_{\Lambda,\Psi}^\text{FWER}$.

\paragraph{Maximum-vs-minimum correlation (range test):} Note that while the $(\Lambda,\Psi)$ pairs in the \emph{adjacent} and \emph{extreme} test are pre-specified, the pair in the \emph{min/max} test is unknown in advance. Therefore in $\Delta_{\Lambda,\Psi}$ we use the two subclasses producing the largest observed delta (by definition of \emph{min/max} test) and use $\overline{\Delta_{\gamma,\psi}}$ as specified above -- the sub-demographics producing the minimal and maximal alignment in each permutation. $p^\text{range}$ denotes that $p$-value obtained for the range text.

\begin{examplebox}{{Worked Example -- Multi-Category Case: Age $c=7$}}

\paragraph{Setup.} The Age attribute has $c = 7$ ordered levels. Group sizes and observed Spearman correlations with GPT-5.2 are:\\

\begin{center}
\begin{tabular}{cccc}
\toprule
Level $c$ & Age group ($\Gamma_c$) & $|U^{\Gamma_c}|$ & $\rho_{\Gamma_c}$ \\
\midrule
1 & 18--24 & 283 & 0.7651 \\
2 & 25--34 & 498 & 0.7852 \\
3 & 35--44 & 586 & 0.7895 \\
4 & 45--54 & 494 & 0.7850 \\
5 & 55--64 & 521 & 0.7635 \\
6 & 65--74 & 438 & 0.7511 \\
7 & 75+    & 191 & 0.7105 \\
\bottomrule
\end{tabular}
\end{center}

\paragraph{One permutation step (illustration).} Consider Split~1, which has 517 respondents. Their age-group labels form a vector of 517 values in $\{1,\dots,7\}$ with fixed per-level counts (e.g., 91 respondents in 35--44 and 26 in 75+ within Split~1). $\pi$ draws a uniformly random permutation of these 517 labels --- keeping the counts for each of the 7 age groups exactly as observed --- and assigns the shuffled labels back, keeping the same number of age labels as in each original split ($|\overline{U^{s,\gamma}}| = |U^{s,\gamma}|$). We recompute all 7 yes-rate vectors and 7 Spearman correlations $\rho_{\Gamma_c}$.

\paragraph{Named pair family} For $c = 7$ we have eight tests: six adjacent pairs $(\Lambda, \Psi) \in (1,2), (2,3), (3,4), (4,5), (5,6), (6,7)$ plus the extreme pair $(\Lambda, \Psi) = (1,7)$. We use $p^\text{FWER}_{\Lambda, \Psi}$ to control for sub-group dependencies setting the null statistic to the $\lambda, \psi$ that yield the max $\overline{\Delta_{\lambda,\psi}}$ in every permutation.

\end{examplebox}

\begin{examplebox}{{Worked Example -- Multi-Category Case: Age $c=7$ \hfill (Cont.)}}

\paragraph{Results after $z = 10^{5}$ permutations:}
\begin{center}
\begin{tabular}{llrrrl}
\toprule
Role & Comparison & $\Delta\rho$ & $p^{\mathrm{raw}}$ & $p^{\mathrm{FWER}}$ & Sig. \\
\midrule
adjacent & 18--24 vs 25--34 & $-0.0201$ & 0.134  & 0.543  & ns \\
adjacent & 25--34 vs 35--44 & $-0.0043$ & 0.643  & 1.000  & ns \\
adjacent & 35--44 vs 45--54 & $+0.0045$ & 0.627  & 0.999  & ns \\
adjacent & 45--54 vs 55--64 & $+0.0215$ & 0.024  & 0.476  & ns \\
adjacent & 55--64 vs 65--74 & $+0.0124$ & 0.214  & 0.901  & ns \\
adjacent & 65--74 vs 75+    & $+0.0406$ & 0.023  & 0.033  & * \\
extreme  & 18--24 vs 75+    & $+0.0545$ & 0.0008 & 0.002  & ** \\
\bottomrule
\end{tabular}
\end{center}

\paragraph{Range test:} The observed maximum correlation is $\rho_{\Gamma_3} = 0.7895$ (35--44 yr) and the minimum is $\rho_{\Gamma_7} = 0.7105$ (75+, see Table above), giving $\Delta_{3,7} = 0.7895 - 0.7105 = 0.0790$. This is larger than the pre-specified extreme pair gap $\Delta_{1,7} = 0.0545$, reflecting that 35--44 year-olds, not the youngest cohort, are most aligned with GPT-5.2. The range null $\overline{\Delta_{\lambda,\psi}}$ automatically accounts for the data-driven selection of this pair, yielding
\begin{equation*}
  p^{\mathrm{range}} = \frac{2}{100,000} \approx 2\times 10^{-5} = (\text{***}).
\end{equation*}
The age gradient in LLM alignment is highly significant, with the oldest respondents (75+) being the least aligned.

\end{examplebox}


\subsection{Topic-Specific Attributes (Topic-Specific Knowledge, View Intensity)}
\label{sec:perm-topic}

Topic-specific knowledge and View intensity are reported separately for each of the three topics (conflicts) --- Russia--Ukraine, Israel--Palestine, and Trump--Harris --- because each respondent reports a separate level for each topic (e.g., a respondent may be ``Knows a great deal'' on Russia--Ukraine while being ``Heard, knows nothing'' on Trump--Harris). For these attributes, level membership is therefore a function of both the respondent and the topic of the question.

Let $\mathcal{T} = \{\text{Rus--Ukr, Isr--Pal, Tru--Har}\}$ index topics, let $\ell_t(u) \in \{1,\dots,c\}$ denote respondent $u$'s level on topic $t \in \mathcal{T}$, and let $Q^t \subseteq Q$ be the questions belonging to topic $t$, with $\bigcup_{t} Q^t = Q$; and $|Q^t| = 144$ ($= Q/3$).

\paragraph{Per-topic pools and observed correlations}
For each topic $t$ and level $c$, the corresponding subgroup is $U^{t,c} = \{u \in U : \ell_t(u) = c\}$. 
the level-$c$ yes-rate on question $q \in \mathcal{Q}^t$ is computed only from respondents whose level on the question's topic equals $c$, so $r_i^c = \frac{y^c_i}{m^c_i}$, where $m^c_i= |U^{tc}|$ -- the total number of responders with level $c$ on topic $t$, and $y^c_i$ is the number of responders in $U^{tc}$ answering positively in question $q_i \in Q^t$. 
Concatenating over topics, we have $R^c$ of the full dimension $k$ that could be correlated with $R^M$.

\paragraph{Per-topic joint shuffle}
Within each topic $t$ \emph{independently} and within each Split $s$ inside that topic, we permute the level labels $\ell_t$ among respondents at any of the $c$ levels, preserving the per-Split per-sub-demographic (/level) counts as specified in the stratified permutation function in Equation \ref{eq:statified_permutation}. The independence across topics reflects the fact that each respondent's three topic-specific levels are separate observations and may be relabeled independently under the null.

Adjacent, extreme, and range statistics, and the corresponding $p$-values $p^{\mathrm{raw}}, p^{\mathrm{FWER}}, p^{\mathrm{range}}$, are the same as defined above.

\subsection{Summary of Results}

Tables \ref{tab:perm-results.1} and \ref{tab:perm-results_knowledge_awareness_stance} report statistical significance for the alignments differences between pairs of demographic subgroups and the GPT-5.2 model. Significance scores are based on all questions in $Q$ (432) The number of permutations per test is $z = 10^{5}$. The family of \emph{Adjacent} and \emph{extreme} pairs are reported through $p^{\mathrm{FWER}}$ (max-statistic correction within Family of significance tests); \emph{max/min} use $p^{\mathrm{range}}$ directly (---  in the $p^{FWER}$ column). Significance codes: *** $p < 0.001$, ** $p < 0.01$, * $p < 0.05$, ns = not significant.

\begin{table}[htbp]
\centering
\caption{Significance tests for various demographic sub-groups ($z = 10^{5}$, M = GPT-5.2, $|Q| = 432$ questions)}
\label{tab:perm-results.1}
\small
\setlength{\tabcolsep}{4pt}
\begin{tabular}{llp{2.5cm}rrrrrl}
\toprule
Attribute & Role & Comparison & $n_A$ & $n_B$ & $\Delta\rho$ & $p^{\mathrm{raw}}$ & $p^{\mathrm{FWER}}$ & Sig. \\
\midrule
Gender         & single & Female vs Male           & 1{,}769 & 1{,}242 & $-0.0007$ & 0.902 & 0.902 & ns \\
\midrule
First language & single & English vs Non-English   & 2{,}720 &   291 & $+0.0411$ & 0.006 & 0.006 & ** \\
\midrule
\multirow{4}{*}{Education}
  & adjacent   & No qual.\ vs Secondary           &   101 & 859 & $-0.0830$ & 0.012 & 0.013 & * \\
  & adjacent   & Secondary vs Undergraduate       &   859 & 775 & $-0.0092$ & 0.213 & 0.994 & ns \\
  & adjacent   & Undergraduate vs Masters/PhD     &   775 & 397 & $+0.0057$ & 0.609 & 0.999 & ns \\
  & extreme    & No qual.\ vs Masters/PhD         &   101 & 397 & $-0.0865$ & 0.004 & 0.008 & ** \\
  \hdashline
Education      & \emph{min/max} & Undergraduate vs No qual.   &   775 & 101 & $+0.0922$ & 0.004 & --- & ** \\
\midrule
\multirow{6}{*}{Social grade}
  & adjacent   & A vs B   & 422 & 625 & $-0.0137$ & 0.192 & 0.730 & ns \\
  & adjacent   & B vs C1  & 625 & 917 & $+0.0054$ & 0.516 & 0.995 & ns \\
  & adjacent   & C1 vs C2 & 917 & 449 & $+0.0041$ & 0.702 & 0.998 & ns \\
  & adjacent   & C2 vs D  & 449 & 311 & $+0.0239$ & 0.051 & 0.179 & ns \\
  & adjacent   & D vs E   & 311 & 287 & $+0.0029$ & 0.821 & 1.000 & ns \\
  & extreme    & A vs E   & 422 & 287 & $+0.0225$ & 0.079 & 0.231 & ns \\ 
  \hdashline
Social grade   & \emph{min/max} & B vs E & 625 & 287 & $+0.0362$ & 0.034 & --- & * \\
\midrule
\multirow{7}{*}{Age}
  & adjacent   & 18--24 vs 25--34 & 283 & 498 & $-0.0201$ & 0.134  & 0.543 & ns \\
  & adjacent   & 25--34 vs 35--44 & 498 & 586 & $-0.0043$ & 0.643  & 1.000 & ns \\
  & adjacent   & 35--44 vs 45--54 & 586 & 494 & $+0.0045$ & 0.627  & 0.999 & ns \\
  & adjacent   & 45--54 vs 55--64 & 494 & 521 & $+0.0215$ & 0.024  & 0.476 & ns \\
  & adjacent   & 55--64 vs 65--74 & 521 & 438 & $+0.0124$ & 0.214  & 0.901 & ns \\
  & adjacent   & 65--74 vs 75+    & 438 & 191 & $+0.0406$ & 0.023  & 0.033 & * \\
  & extreme    & 18--24 vs 75+    & 283 & 191 & $+0.0545$ & 0.0008 & 0.002 & ** \\
  \hdashline
Age            & \emph{min/max} & 35--44 vs 75+ & 586 & 191 & $+0.0790$ & 3\,$\times$\,10$^{-5}$ & --- & *** \\
\bottomrule
\end{tabular}
\end{table}

\begin{table}[htbp]
\centering
\caption{Significance tests for Topic-Awareness and View intensity levels ($z = 10^{5}$, M = GPT-5.2, $|Q| = 432$ questions). }
\label{tab:perm-results_knowledge_awareness_stance}
\small
\setlength{\tabcolsep}{4pt}
\begin{tabular}{llp{2.5cm}rrrrrl}
\toprule
Attribute & Role & Comparison & $n_A$ & $n_B$ & $\Delta\rho$ & $p^{\mathrm{raw}}$ & $p^{\mathrm{FWER}}$ & Sig. \\
\midrule
\multirow{3}{*}{Gen.\ political awareness}
  & adjacent & Low (0--3) vs Medium (4--6)      &   520 &   966 & $-0.0082$ & 0.386 & 0.606 & ns \\
  & adjacent & Medium (4--6) vs High (7--10)    &   966 & 1{,}525 & $-0.0238$ & 0.0002 & 0.018 & * \\
  & extreme  & Low (0--3) vs High (7--10)       &   520 & 1{,}525 & $-0.0320$ & 0.0004 & 0.0007 & *** \\
  \hdashline
Gen.\ political awareness & \emph{min/max} & High (7--10) vs Low (0--3) & 1{,}525 & 520 & $+0.0320$ & 0.0007 & --- & *** \\
\midrule
\multirow{3}{*}{Topic-specific knowledge}
  & adjacent & Heard, knows nothing vs Knows something   &   294 & 2{,}053 & $-0.0671$ & $10^{-5}$ & $10^{-5}$ & *** \\
  & adjacent & Knows something vs Knows a great deal     & 2{,}053 &   649 & $+0.0020$ & 0.826 & 0.993 & ns \\
  & extreme  & Heard, knows nothing vs Knows a great deal &  294 &   649 & $-0.0651$ & $10^{-5}$ & $10^{-5}$ & *** \\
  \hdashline
Topic-specific knowledge & \emph{min/max} & Knows something vs Heard, knows nothing & 2{,}053 & 294 & $+0.0671$ & $10^{-5}$ & --- & *** \\
\midrule
\multirow{3}{*}{View intensity}
  & adjacent & No strong views vs Slight lean    &   858 &   739 & $-0.0304$ & 0.0002 & 0.0004 & *** \\
  & adjacent & Slight lean vs Strong lean        &   739 & 1{,}414 & $+0.0038$ & 0.619 & 0.882 & ns \\
  & extreme  & No strong views vs Strong lean    &   858 & 1{,}414 & $-0.0266$ & 0.0008 & 0.002 & ** \\
  \hdashline
View intensity  & \emph{min/max} & Slight lean vs No strong views &   739 &   858 & $+0.0304$ & 0.0004 & --- & *** \\
\bottomrule
\end{tabular}
\end{table}

\section{Alignment across Topics}
\label{sm:Alignment_across_Topics}

\begin{figure}[H]
    \centering
    \includegraphics[width=0.5\linewidth]{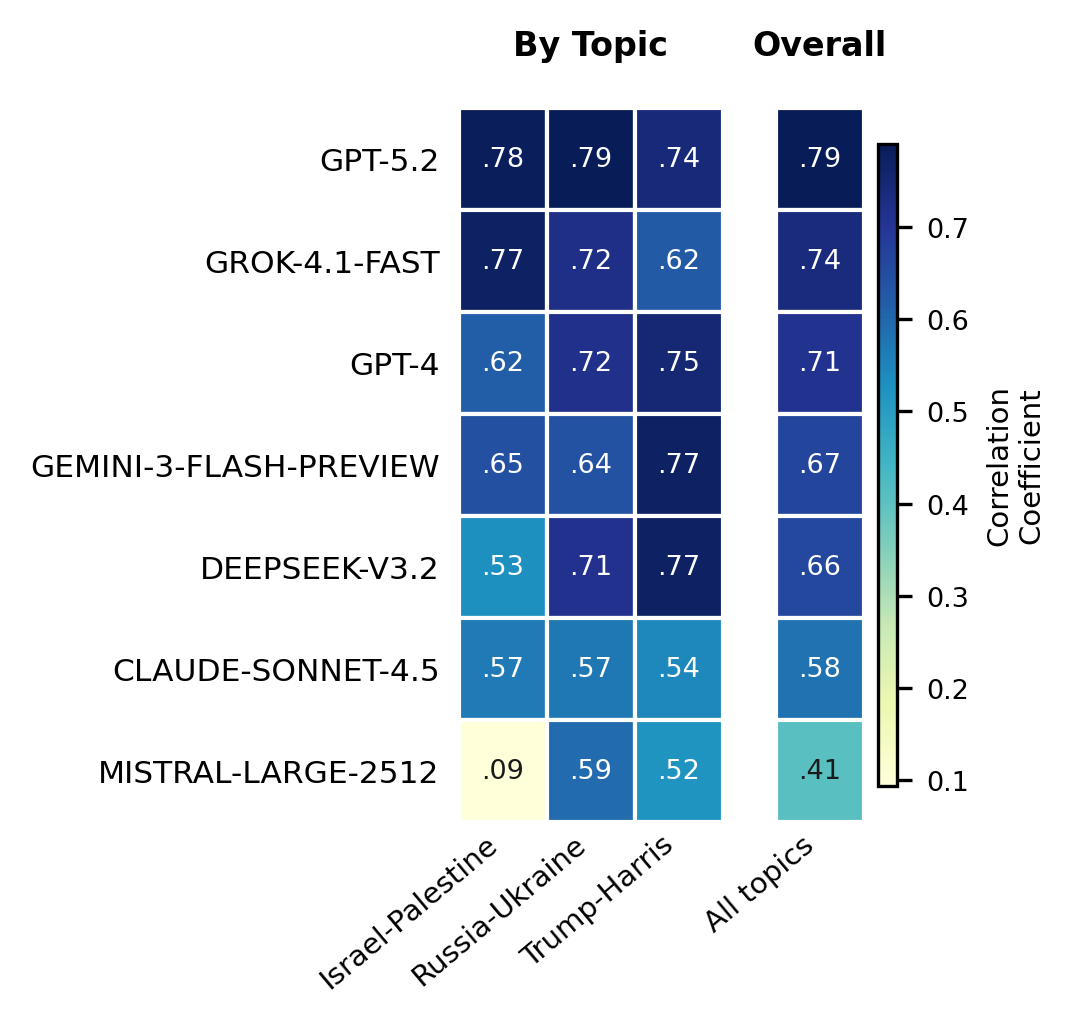}
    \caption{Models' alignment score per each of the three different topics} 
    \label{fig:Alignment across Topics}
\end{figure}

Mistral's markedly lower alignment ($\rho = 0.41$) is not an artifact of unparsable or refused responses - coercion rates were negligible across all models. Rather, it reflects two compounding tendencies. First, Mistral answered almost deterministically, pinning 98\% of its question-level scores to 0 or 1 (versus 11\% of human responses), which collapses the graded signal needed to track human sympathy proportions. Second, and more decisively, its weakness was concentrated almost entirely in a single topic: while it achieved moderate alignment on the Russo-Ukrainian war ($\rho = 0.59$) and the Trump–Harris campaign ($\rho = 0.52$), its alignment on the Israel–Palestine headlines was effectively random ($\rho = 0.09$, versus 0.78 for GPT-5.2). On these items Mistral treated Israel-sympathy as a near coin-flip uncorrelated with human readers and systematically under-registered Gaza sympathy, defaulting to "No" on 75\% of headlines even though human respondents leaned sympathetic. Thus Mistral's poor aggregate performance masks a topic-specific collapse rather than uniformly weak comprehension - demonstrating how models' alignment could differ per topic, not only demographic variables. Because this phenomenon was largely unique to Mistral in our dataset - at least in its magnitude - we report the pooled results in the main text to preserve readability and interpretability, while providing topic-specific analyses in the supplemental material (see Fig.~\ref{fig:Alignment across Topics}).

\bibliographystyle{naturemag} 
\bibliography{llm_representation} 